\documentclass[runningheads]{llncs}
\usepackage{todonotes}
\usepackage{graphicx}
\usepackage{booktabs}

\usepackage{multirow}
\usepackage{array}
\usepackage{siunitx}
\usepackage{dblfloatfix}
\usepackage{subfigure}

\newcolumntype{R}[1]{>{\raggedleft\let\newline\\\arraybackslash\hspace{0pt}}m{#1}}
\newcolumntype{L}[1]{>{\raggedright\let\newline\\\arraybackslash\hspace{0pt}}m{#1}}
\newcolumntype{C}[1]{>{\centering\let\newline\\\arraybackslash\hspace{0pt}}m{#1}}

\begin{document}
\title{Investigating Correlations of Inter-coder Agreement and Machine Annotation Performance for Historical Video Data}
\titlerunning{Correlations of Inter-coder Agreement and Machine Annotation Performance}


\author{Kader Pustu-Iren\inst{1,2} \and
Markus M{\"{u}}hling\inst{3}
\and Nikolaus Korfhage\inst{3}
\and Joanna Bars\inst{4}
\and Sabrina Bernh{\"{o}}ft\inst{4}
\and Angelika H{\"{o}}rth\inst{4}
\and Bernd Freisleben\inst{3}
\and Ralph Ewerth\inst{1,2}}
\authorrunning{K. Pustu-Iren et al.}
\institute{Leibniz Information Centre for Science and Technology (TIB), Hannover, Germany \and
L3S Research Center, Hannover, Germany \\
\email{\{kader.pustu,ralph.ewerth\}@tib.eu} \and
Department of Mathematics and Computer Science, University of Marburg, Marburg, Germany\\
\email{\{muehling,korfhage,freisleb\}@informatik.uni-marburg.de} \and
German Broadcasting Archive (DRA), Potsdam, Germany\\
\email{\{joanna.bars,sabrina.bernhoeft,angelika.hoerth\}@dra.de}}
\maketitle              
\begin{abstract}
Video indexing approaches such as visual concept classification and person recognition are essential to enable fine-grained semantic search in large-scale video archives such as the historical video collection of former German Democratic Republic (GDR) maintained by the German Broadcasting Archive (DRA). 
Typically, a lexicon of visual concepts has to be defined for semantic search. However, the definition of visual concepts can be more or less subjective due to individually differing judgments of annotators, which may have an impact on annotation quality and subsequently training of supervised machine learning methods. In this paper, we analyze the inter-coder agreement for historical TV data of the former GDR for visual concept classification and person recognition. The inter-coder agreement is evaluated for a group of expert as well as non-expert annotators in order to determine differences in annotation homogeneity.
Furthermore, correlations between visual recognition performance and inter-annotator agreement are measured. In this context, information about image quantity and agreement are used to predict average precision for concept classification. Finally, the influence of expert vs. non-expert annotations acquired in the study are used to evaluate person recognition.
\end{abstract}
\section{Introduction}
\begin{figure}[t]
  \centering
  \includegraphics[width=0.95\linewidth]{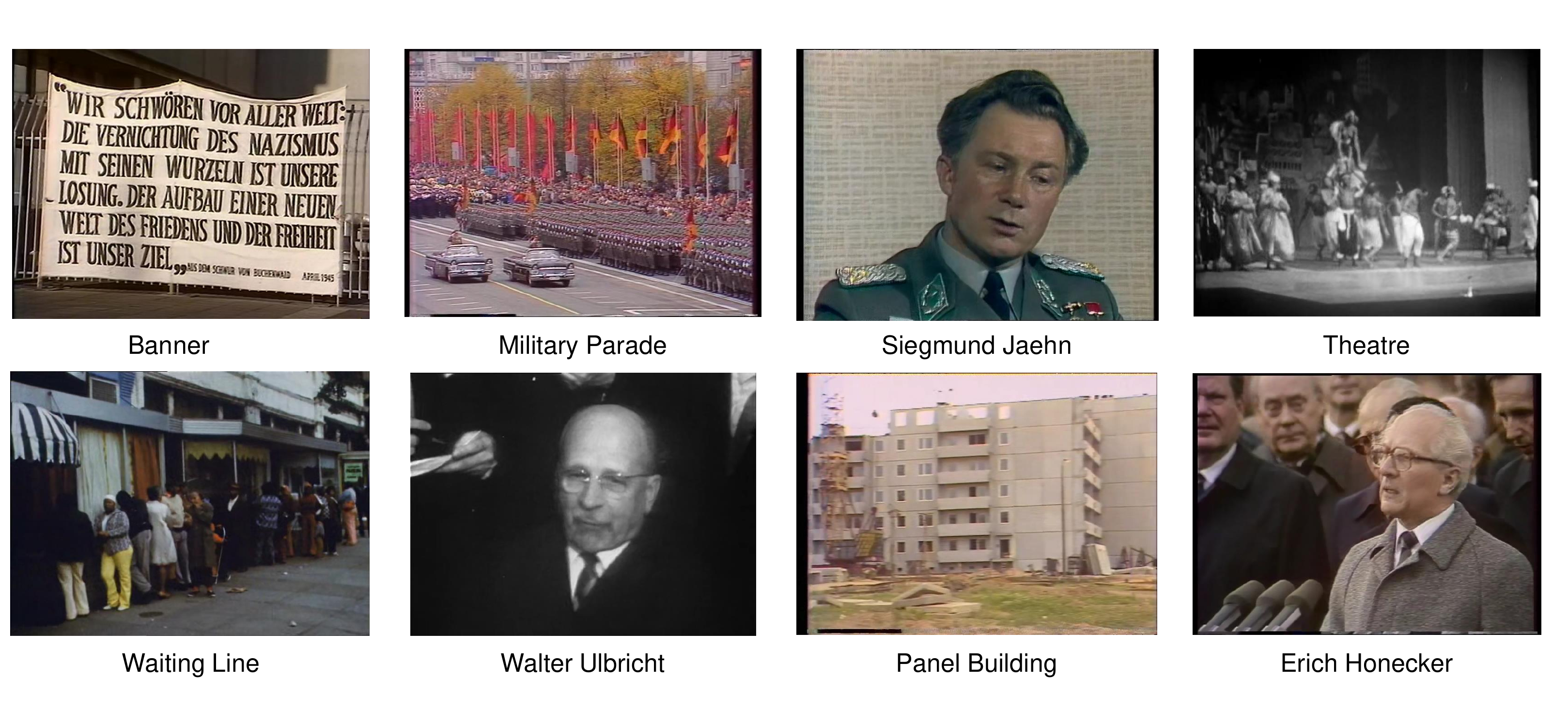}
  \vspace{-0.5cm}
  \caption{Example category samples of our inter-coder annotation study.}
  \label{fig:intro_images}
\end{figure}

Automatic indexing is an important prerequisite to enable semantic search in large video archives. In particular, visual concept classification and person recognition play an essential role to provide fine-grained access to large image and video databases like the historical video collection of the German Broadcasting Archive (DRA). The DRA maintains the cultural heritage of television broadcasts of the former German Democratic Republic (GDR) and thus grants access to researchers who are interested in German-German history. Semantic search through pre-defined lexicons of visual concepts and personalities associated with the former GDR can support the investigation of specific research questions and be a starting point for further analyses and scientific studies. Typically, a lexicon of visual concepts and persons has to be defined in advance. However, in order to allow for such a fine-grained search by automatically indexing the video collection, a huge manual effort for initializing this process is necessary. A large amount of manually labeled keyframes and shots is necessary to train or fine-tune deep learning based video indexing approaches. While the number of images is known to matter for these approaches, the quality of manually annotated training data affects video indexing performance. In this context, visual concepts in images and video frames are not always perceived objectively. As shown in previous work~\cite{Nowak2010,Ewerth2017}, human annotators can have different  understandings and  judgments for certain concepts and the inter-coder agreement noticeably vary for different visual concepts. Therefore, the precise definition of such concepts and consequently carefully labelled data are crucial for system success.

In this paper, we investigate the inter-coder agreement for annotations of historical TV data of the former GDR for expert and non-experts. The quality of manually labeled keyframes for both groups is assessed for  visual concept classification and person recognition.
Moreover, correlations between inter-coder agreement and system performance by means of average precision results on the two tasks of visual concept classification and person recognition are computed and analyzed. Our hypothesis is that, besides the amount of available training data, the inter-coder agreement is correlated to the performance of video indexing methods based on supervised learning.
Moreover, we suppose that inter-coder agreement might form an upper bound for such methods. In this regard, some first experiments are presented to predict average precision using support vector regression based on inter-coder agreement and training data size.
Furthermore, the influence of image annotation quality induced by experts vs. non-experts on the person recognition performance is investigated based on annotations acquired during the study.

The remainder of the paper is organized as follows. Section \ref{relwork} discusses related work regarding inter-annotator studies. Section \ref{sec:agreement} deals with the comprehensive user study including a description of the used dataset, the study participants, the experimental design and the results of a comparison between expert vs. non-expert inter-coder agreements. Furthermore, a performance prediction for the task of concept classification is presented based on inter-coder agreement and training data size as input. In Section \ref{sec:personid}, the impact of expert vs. non-expert image annotations on person identification performance is investigated. Section \ref{conclusions} concludes the paper and outlines areas for future work.
\section{Related Work}
\label{relwork}
In this section, we briefly survey related work for inter-annotator studies conducted for natural language as well as image annotation tasks. Snow et al.~\cite{Snow2008} have evaluated non-expert annotations by means of Mechanical Turk workers for natural language tasks. Among other experiments, they found that annotations of four non-experts are needed to rival the annotation quality of one expert annotator in selected tasks. Furthermore, they have trained machine learning classifiers on expert as well as non-expert annotations and reported better system performance for non-experts due to high annotation diversity reducing the annotator bias. However, these observations were exclusively made for natural language tasks.
Nowak and Rueger~\cite{Nowak2010} presented a study on inter-coder agreement for image annotation from both crowdsourcing and experts. Human annotators had to label 99 images of the ImageCLEF test data (\url{http://www.imageclef.org/}) with respect to 22 concept categories. Some of the categories were mutually exclusive (season, time of day, indoor/outdoor/none). The images were assessed by experts as well as by Mechanical Turk workers. They measure higher agreement for experts, but argue that majority voting filters out noise in non-expert annotations closing the gap to expert annotations of higher quality. A more recent study~\cite{Ewerth2017} deals with the question whether machines perform better than humans in visual recognition tasks. For assessing human performance the inter-coder reliability by Krippendorff's $\alpha$ on 20 common categories of the PASCAL VOC benchmark is measured. For the best submission at PASCAL VOC's leaderboard an above average human-level performance for visual concept annotation is reported being on a par or better than 19 of 23 participants.
\section{Annotation Study: Expert vs. Non-Expert Agreement on GDR-specific Concepts and Persons}
\label{sec:agreement}
In this section, we aim at comparing the reliability of expert vs. non-expert annotations for for historical TV data in terms of inter-coder-agreement. For this purpose, we collected annotations of expert as well as student participants (Section~\ref{participants}) on a selected set of concept and person images~(Section~\ref{dataset}) following the experimental design described in Section~\ref{experiments}. Agreement results are discussed in Section~\ref{agreement_results}.
\subsection{Dataset of Historical Concepts and Personalities}
\label{dataset}
\begin{table}[!bp]
\caption{Concepts and persons of our study along with the number of selected study image samples $|S|$ and former Top-200 AP results per category. Additionally, also the number of training images $|T|$ used to train the concept and person recognition models are reported. }
\label{tab:classes}
\begin{minipage}{.48\linewidth}
  \centering
  \begin{tabular}{L{3.45cm} R{0.8cm} C{0.8cm} C{0.8cm}}
  \toprule
  Concept & $|T|$ & $|S|$ & AP\\ 
  \midrule
    Apartment construction & 2,753 & 54 & 0.78 \\
    Automotive industry    & 617 & 30 & 0.27 \\
    Banner                 & 2,581 & 62 & 0.92 \\
    Camping                & 346 & 28 & 0.56 \\
    Kindergarten           & 1,854 & 29 & 0.35 \\
    Microelectronics       & 1,281 & 15 & 0.13 \\
    Military parade        & 2,809 & 67 & 0.96 \\
    Mining                 & 1,885 & 15 & 0.20 \\
    Narrow-gauge railway   & 444 & 10 & 0.12 \\
    Open-pit mining        & 2,759 & 61 & 0.78 \\
    Panel building         & 3,307 & 66 & 0.98 \\
    Playground             & 632 & 34 & 0.48 \\
    Prison                 & 402 & 10 & 0.07 \\
    Ship launching         & 364 & 24 & 0.27 \\
    Shipyard               & 1,079 & 35 & 0.22 \\
    Shopping mall          & 1,622 & 55 & 0.66 \\
    Textile factory        & 2,209 & 60 & 0.77 \\
    Theater performance    & 1,172 & 20 & 0.19 \\
    Unrefurbished house    & 835 & 59 & 0.59 \\
    Waiting line           & 313 & 10 & 0.01 \\
    \bottomrule
  \end{tabular}
  \label{tab:filtering_results}
\end{minipage}%
\begin{minipage}{.04\linewidth}
  \quad
\end{minipage}%
\begin{minipage}{.47\linewidth}
  \centering
	\label{tab:domain_info}
	\begin{tabular}{L{3.15cm} R{0.8cm} C{0.8cm} C{0.8cm}}
	\toprule
	Person & $|T|$ & $|S|$ & AP\\
	\midrule
    Christa Wolf       & 30 & 11 & 0.76 \\
    Erich Honecker     & 1,171 & 63 & 1.00 \\
    Fritz Cremer       & 17 & 13 & 0.62 \\
    Hermann Henselmann & 8 & 10 & 0.86 \\
    Hilde Benjamin     & 28 & 14 & 0.96 \\
    Siegmund Jaehn      & 31 &10 & 0.98 \\
    Stephan Hermlin    & 29 & 10 & 0.47 \\
    Walter Ulbricht    & 256 & 28 & 1.00 \\
    Werner Tuebke       & 21 & 15 & 0.65 \\
    Mikhail Gorbachev  &  - & 25 & -   \\
    \bottomrule
  \end{tabular}
\end{minipage}
\end{table}
In a previous project that realized content-based video retrieval in historical GDR television data, a GDR specific lexicon of 100 classes (91 concepts and 9 persons) has been defined based on the utility and usefulness for anticipated search queries~\cite{Muehling2016}. The concept classification was based on a multi-label CNN approach using a GoogleNet architecture~\cite{Szegedy2015}. Person recognition was performed using FaceVACs~\footnote{\url{http://www.cognitec.com}}, a commercial approach combining face alignment and recognition. Visual Concept and person recognition models have been applied to a test set of about 2,500 hours of video data and evaluated on the Top-200 retrieval results in terms of mean average precision. Therefore, the Top-200 retrieval results per class had been manually labeled by an expert~\cite{Muehling2016}. For the current study, 20 concepts and 9 persons have been chosen from the GDR-specific lexicon. The concepts are selected based on the average precision performance so that concepts with high, middle and low retrieval quality are evenly represented. The images for the study have been randomly sampled from the Top-200 retrieval results of the selected concepts and persons. Altogether, 744 positive images for concepts and 199 positive images of persons are incorporated. This includes additional images for rarely occurring classes and an extra person ("Mikhail Gorbachev").
To ensure a minimum of 10 images per class, the additional images are collected from the test set using metadata information and video OCR (optical character recognition. To further enlarge the dataset, 257 images with negative labels are randomly chosen from the Top-200 retrieval results (178 images from concept retrieval results and 79 images from person retrieval results). These images of unknown content tend to be similar to one of the concepts or persons, making the manual labeling process more difficult. Finally, the image annotation study comprises 1200 images, 20 concepts and 10 persons as presented in Table~\ref{tab:classes}.
\subsection{Study Participants}
\label{participants}
Five experts and five non-experts participated in our inter-coder agreement study. The group of experts consists of DRA employees, who are in a general sense very familiar with the historical material we use in our study. In particular, experts are fully trained archivists and information specialists with a mean age of $40(\pm10)$ years and work experiences ranging from 3 to 28 years. The non-expert group is composed of Master's students in computer science with a mean age of $26(\pm3)$ years, who have not worked with the material previously. For the participation in the study, students were rewarded 30 Euro.
\subsection{Experimental Design}
\label{experiments}
Participants were asked to classify historical images into 20 concept as well as 10 person categories. Thus, we allowed to assign multiple labels, for the case that more than one concept was detected in an image. Since we included random images depicting none of the available categories, participants had the choice to skip and leave images unlabeled.

Participants were provided with annotation instructions which they were allowed to keep during the study. For the concept categories we provided formal descriptions as well as some hints how certain concepts may appear in the images. For each of the personalities we provided three images, which were chosen from both Google and further archive material. The images were selected such that the person's variability in appearance and age was covered as much as possible. Furthermore, we made sure that example images were mutually exclusive with the ones in the study. Before the start of the annotation task, the participants were given the opportunity to look up concepts and person on the Internet in case of uncertainties. However, they were prohibited to search the Internet during the study.

To evaluate the degree of agreement between different sets of annotators, we used Krippendorff's alpha (K's $\alpha$) coefficient~\cite{Krippendorff2004}. We chose K's $\alpha$ instead of other reliability measures, since it is not affected by the number of coders~\cite{Hayes2007}. Hence, reliability can be compared for annotator groups of asymmetric size. Moreover, K's $\alpha$ is able to handle missing data. This also suites our study setup well since we involved images which do not belong to any of our categories and allowed participants to skip images unlabeled.
\subsection{Results}
\label{agreement_results}
\begin{figure}[!bp]
\centering
\begin{subfigure}
\centering
\includegraphics[width=0.85\textwidth]{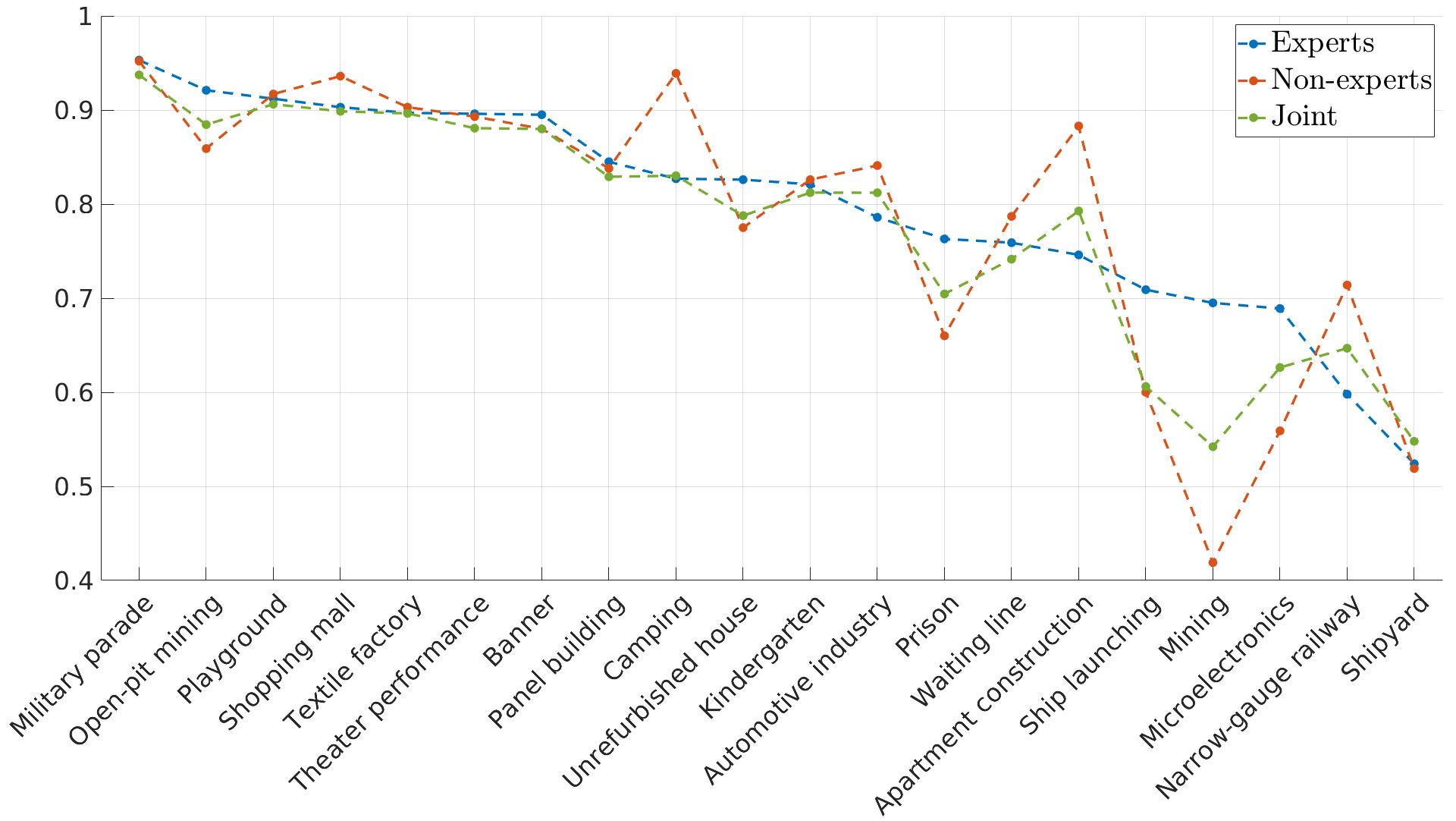}
\end{subfigure}
\begin{subfigure}
\centering
\hspace{0.2cm}
\includegraphics[width=0.85\textwidth]{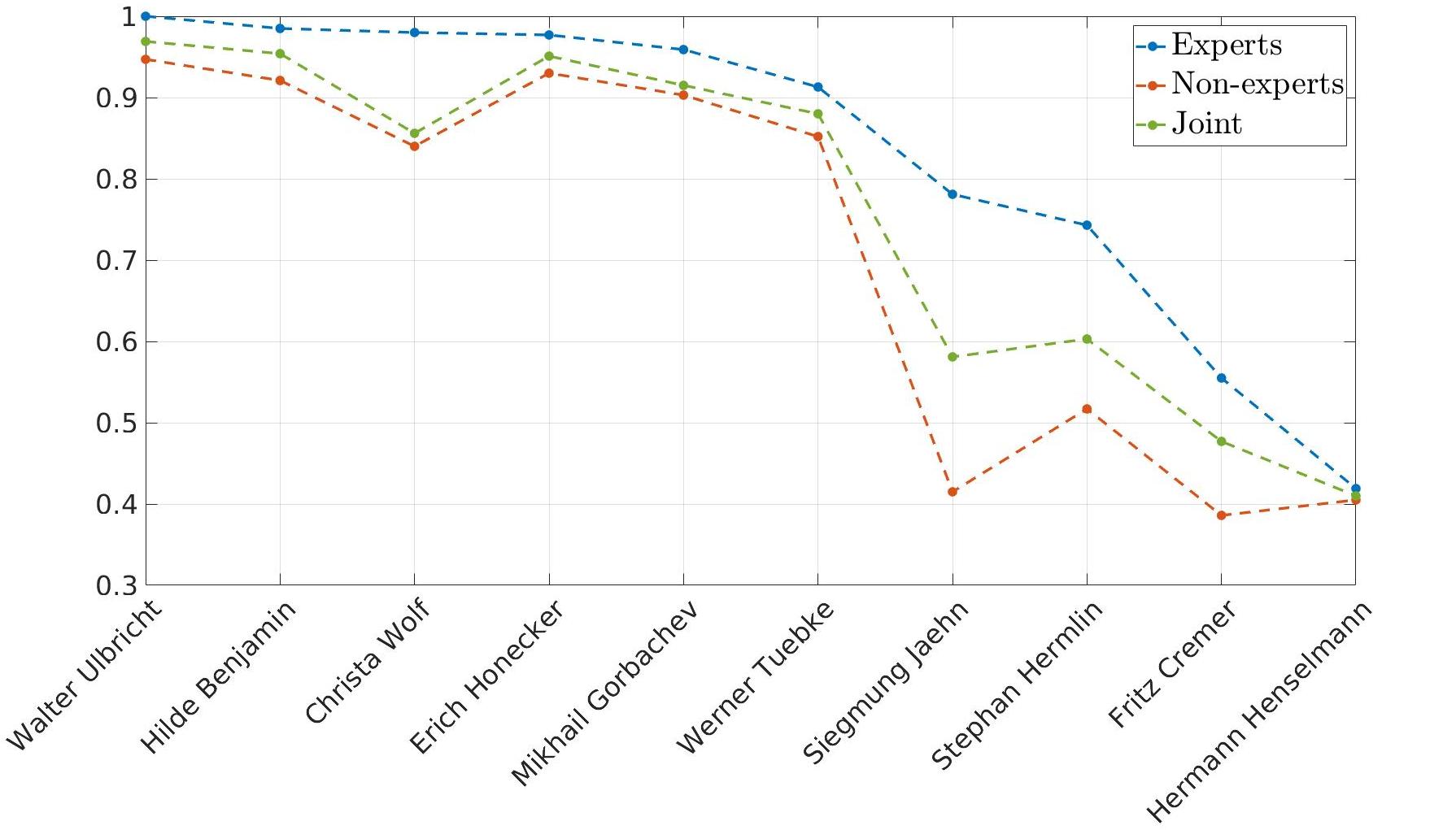}
\end{subfigure}
\caption{Inter-coder agreement of experts vs. non-experts in terms of K's $\alpha$'s for concepts (top) as well as persons (bottom). Additionally, joint agreement including both expert and non-expert annotators is shown.}
\label{fig:plot_agreement}
\end{figure}
The resulting inter-coder agreement in terms of Krippendorff's $\alpha$ for the 1200 image samples of our study is displayed in Figure~\ref{fig:plot_agreement}. Thus, we report K's $\alpha's$ per category, respectively for the group of experts and non-experts.
In order to measure how well non-experts and experts agree on the categories, we also determined combined agreement among both groups. While experts and non-experts agree almost equally strong on the concept categories ($0.80$ vs. $0.79$ across all concepts), expert annotations seem to be more homogeneous for the GDR personalities ($0.83$ vs. $0.71$ across all personalities).
According to Krippendorff~\cite{Krippendorff2004}, an $\alpha > 0.8$ can be considered an agreement value for reliable annotations. Therefore, annotations on concepts can be counted reliable for both groups of experts and students.
Considering single concepts, there are many categories non-experts do surprisingly well on (12 concepts above reference value of $0.8$). In comparison to the experts, they failed stronger on concepts like \emph{mining}, \emph{microelectronics} or \emph{ship launching}. For some of these categories the provided images are darker or low-quality close-ups, which are hard to classify due to either similar categories or lacking knowledge on the historical context. Nevertheless, non-experts were able to annotate 12 concept categories validly, whereas expert annotations of 11 concepts are considered reliable.

For persons, greater differences between experts and non-experts can be observed
suggesting that experts are better annotators for the person identification task. The non-expert agreements are lower for all selected GDR personalities. While non-experts accurately identify commonly known personalities like \emph{Erich Honecker}, \emph{Mikhail Gorbachev} or \emph{Walter Ulbricht}, the largest insecurities can be observed for persons like \emph{Fritz Cremer} or \emph{Hermann Henselmann}. This could be due to images of poor quality and also to a larger variability in appearance for the latter person, making it harder to recognize them. Though overall only the person annotations of the expert group are considered a reliable source according to Krippendorff's rule, non-expert annotations seem to be 'good enough' for the same personalities that experts annotated validly.
%
%

For the agreement of aggregated expert and non-expert annotations also some interesting observations can be made. Overall, an averaged joint agreement of $0.78$ on concepts and $0.76$ on persons can be determined. This implies that experts agree stronger with isolated experts than with the combined group, especially in the case of persons. Non-experts, on the contrary, agree stronger with experts in the case of persons. Surprisingly, the overall concept agreement in the inter-group scenario is also lower than the agreement for the weaker non-expert annotators. This could suggest that the groups of archivists and students may have different understandings for some of the concepts causing diverging biases. Some example concepts for which inter-group agreement is slightly lower than both isolated expert and non-expert agreement are \emph{waiting line}, \emph{military parade} and \emph{theater performance}. 
Superior joint agreement of experts and non-experts in comparison to single-groups agreement can be uniquely observed for the concept \emph{shipyard} due to a set of stronger agreeing experts and students.

Earlier in this section we implied that a high agreement is an indicator for an accurate set of annotations. For the categories of concepts, experts and non-experts were found to be equally accurate annotators.
In the next section, we analyze to which extent agreement contributes to video indexing performance and investigate the opportunity of performance prediction.
We determined that experts are more accurate labelers when it comes to historically relevant persons of the GDR. However, judgments of different groups may have different biases as also reasoned for the concept categories. Therefore, in Section~\ref{sec:personid}, we exemplary determine for the set of different personality annotations collected in our study whether high agreement on training images implies higher system performance for person recognition.
\section{Predicting Concept Classification Performance }
\label{sec:performance_prediction}
In this section, we discuss correlations between original average precision results and inter-coder agreements from our study as well as training data size (Section~\ref{sec:correlations}). Afterwards, we exploit determined correlations in order to predict AP (average precision) of visual concept classification depending on the amount of training data and the agreement on concepts (Section~\ref{subsec:performance_prediction}).
\subsection{Correlations with Average Precision Results}
\label{sec:correlations}
\begin{figure}[!bp]
  \centering
  \includegraphics[width=0.7\linewidth]{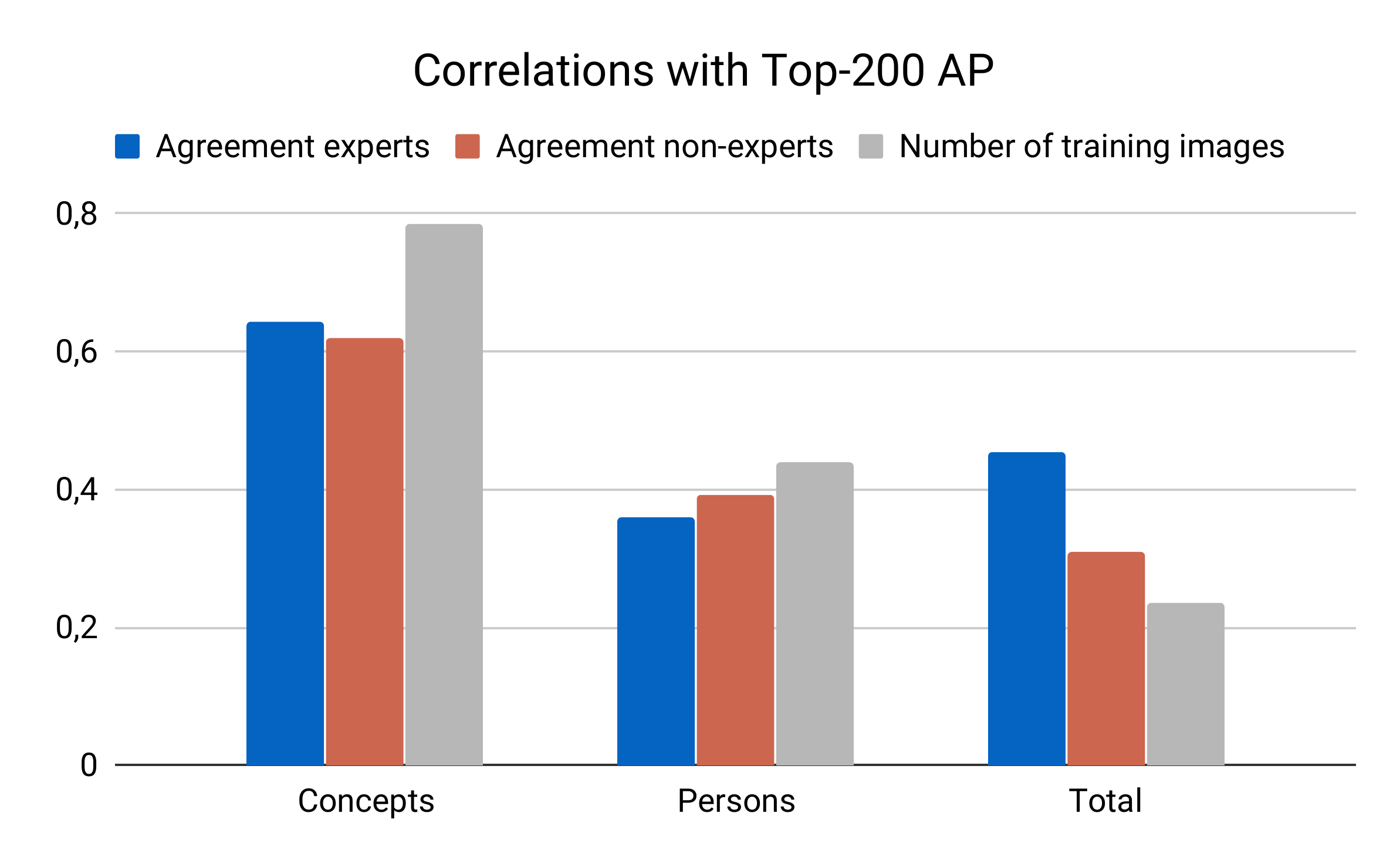}
  \vspace{-0.5cm}
  \caption{Correlations measured between 1) agreement of experts, 2) agreement of non-experts, 3) number of training images and Top-200 AP on concepts, persons and all categories.}
  \label{fig:chart_correlations}
\end{figure}
In Figure~\ref{fig:chart_correlations}, correlations between the original average precision results presented in Table~\ref{tab:classes}, the inter-coder agreement and the number of training images (see also Table~\ref{tab:classes}) are displayed. Thus, the highest correlations are measured for AP results on concepts. While the correlation with expert agreement (0.64) is slightly higher than with non-expert agreement (0.62), also a significant correlation between AP values and the number of training images can be measured (0.78).
We can notice only moderately high correlations for AP on persons as well as for total AP results. For this reason, we exploit high correlations observed for concepts in order to perform a performance prediction based on both inter-coder agreement and training data size for the task of concept classification.
\begin{figure}[!tp]
\hspace{-0.1cm}
  \includegraphics[width=1\linewidth]{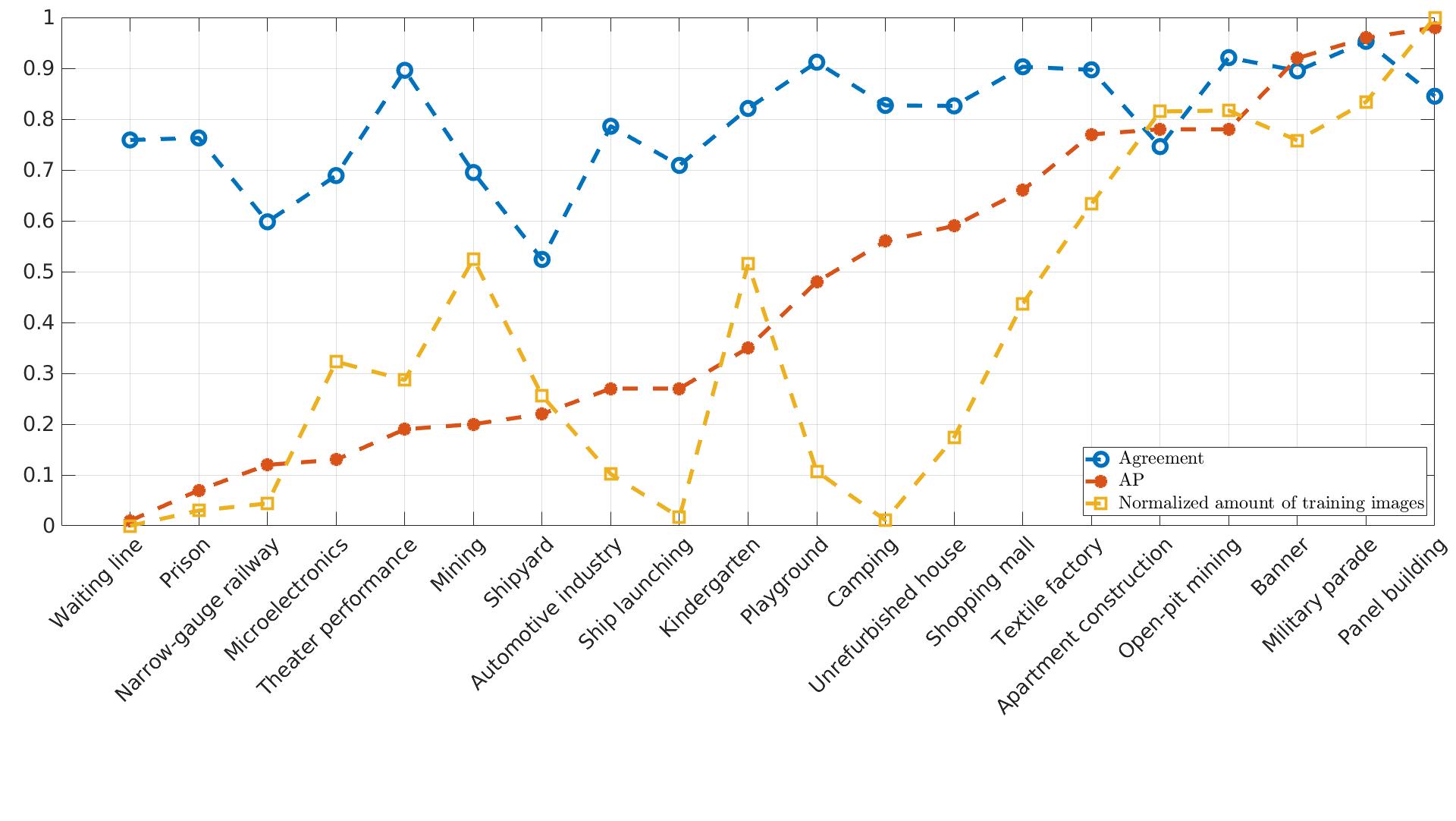}
  \vspace{-1.2cm}
  \caption{Comparison of AP, expert's agreement and normalized training data size.}
  \label{fig:plot_AP_alpha}
\end{figure}
The concepts in Figure~\ref{fig:plot_AP_alpha} are sorted according to AP and show a comparison to inter-coder agreement and normalized training data size. For example, it can be noticed that for nine out of the best 10 concepts, the inter-coder agreement is above 0.8, whereas it is below 0.8 for the other ten concepts. Furthermore, it can be observed that the best six concepts have the highest number of training images.
\subsection{Performance Prediction for Concept Classification}
\label{subsec:performance_prediction}
We employ support vector regression (SVR)~\cite{Smola2004} based on a linear kernel to estimate AP on our study concepts given the corresponding training data sizes and agreement values as input. SVR is performed separately for expert and student agreement. According to leave-one-out cross validation, utilizing expert agreement yields a mean absolute error of $13.95 \%$ ($\pm 12.3 \%$). Using non-expert agreement as input instead, yields a slightly higher error of $14.55 \%$ ($\pm 10.6 \%$).
In order to determine the individual impact of number of training images and annotation quality on predicting the AP on concepts, we also train SVR with training data size and expert agreement as input separately.
Thus, the mean absolute error for using solely training data size as input is $20.0 \%$ ($\pm 10.8 \%$). AP prediction based on the agreement only, yields an error of $21.1 \%$ ($\pm 14.6 \%$). In comparison, the error of a random baseline using prediction by chance is around $33.1 \%$.
\begin{figure}[!tp]
  \includegraphics[width=1\linewidth]{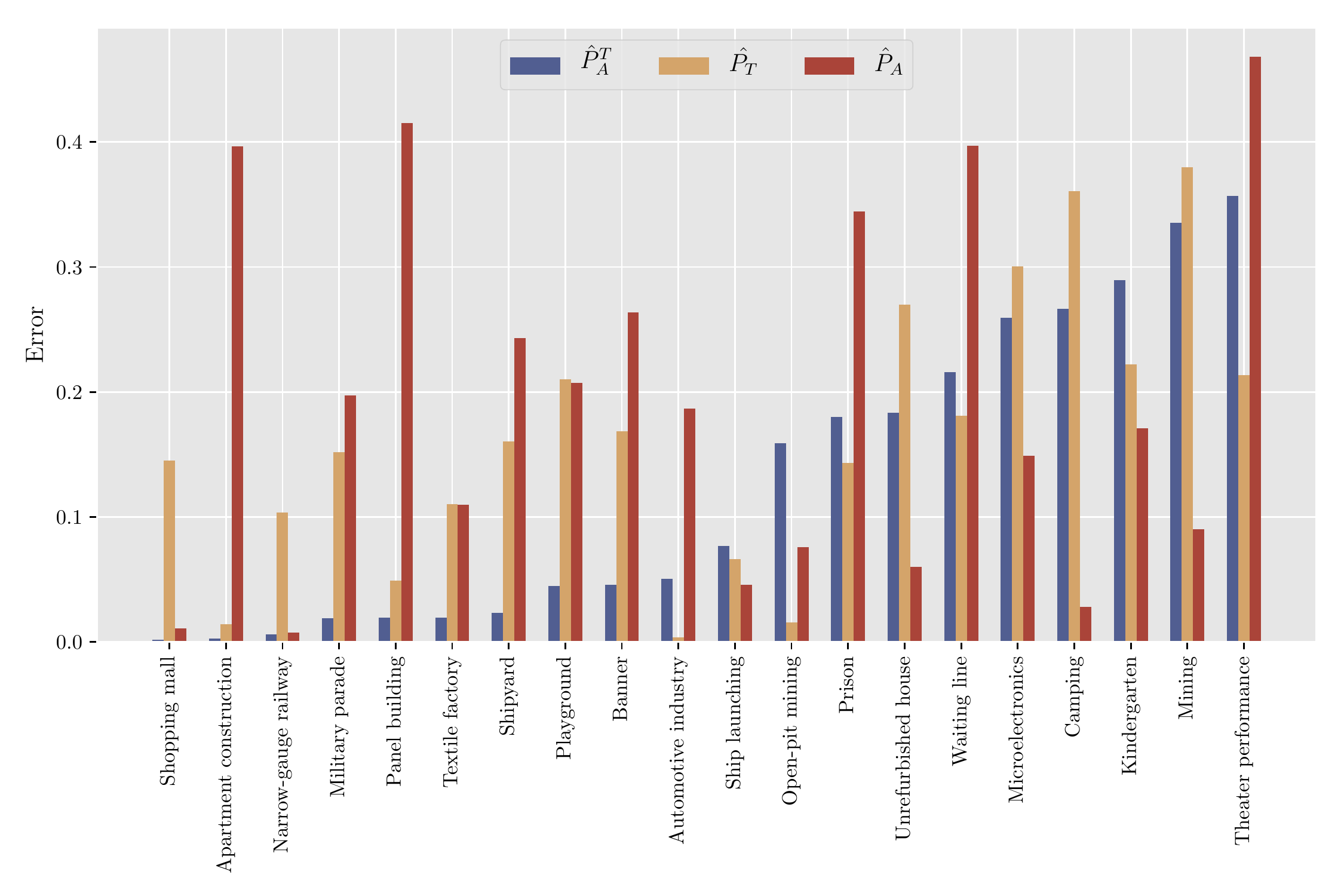}
  \vspace{-0.7cm}
  \caption{Errors of AP estimates on concepts based on prediction with 1) agreement as well as training image size ($\hat{P}^T_A$), solely 2) with training image size ($\hat{P}_T$) and 3) agreement ($\hat{P}_A$)as input.}
  \label{fig:barplot_error}

\end{figure}
Figure~\ref{fig:barplot_error} shows errors on the individual concepts for the different performance estimation approaches. It can be seen that the combined input variables significantly reduce the error for some of the concepts and overall lead to lower errors for most of the concepts. Thus, training image size-based and especially agreement-based prediction lead to high errors for many concepts. Overall, estimation results suggest that the AP performance on the historical concepts can be jointly approximated by training image quantity and image annotation quality. Using expert agreement and performing SVR on all concepts as training samples (i.e., no leave-one-out setting), AP can be \textit{exemplary} approximated by the following function:
\begin{equation}
    \hat{P}^{T}_{A} = 0.379 \cdot \alpha' + 0.698 \cdot t' + 0.018
\end{equation}
where $\alpha'$ corresponds to normalized agreement K's $\alpha$ and $t'$ to the normalized training data size $\in [0,1]$. However, $\hat{P}^{T}_{A}$ have to lie within $[0,1]$ for valid estimates, i.e., results for $\hat{P}^{T}_{A}>1.0$ are set to $1.0$.
\section{Impact of Expert vs. Non-Expert Annotations on Person Identification Performance}
\label{sec:personid}
In this section, we evaluate person identification performance based on different sets of expert and non-expert annotations acquired in the inter-coder annotator study (Section~\ref{sec:agreement}). In particular, we create annotation-dependant classifiers following a face matching approach and determine systems based on single participant's votes as well as majority vote for both experts and non-experts.
\subsection{Experimental Setup}
\subsubsection{Person Identification Method}
For the identification of the introduced ten personalities associated with GDR, we employ a basic face matching approach, in which comparisons rely on deep feature representations. In order to build a dictionary of facial representations, each personality is represented by the mean of their embeddings which are generated from a ground truth set of face images. Thus, an unknown face is considered a person of the dictionary, if the cosine similarity between its embedding and a dictionary embedding is sufficiently high according to a cosine similarity threshold. In order to extract deep facial features, we employ a FaceNet~\cite{schroff2015facenet} implementation~\footnote{\url{https://github.com/davidsandberg/facenet}}, which was trained on the \emph{VGGFace2} dataset~\cite{Cao2018} using the \emph{Inception ResNet v1} architecture~\cite{Szegedy2016}. The model achieves a LFW~\cite{huang2007labeled} accuracy of $99.65$~\%. We use the cosine similarity threshold of $0.67$ from the FaceNet implementation code.

\subsubsection{Training and Test Set}
We use the judgments on the GDR personalities of the different annotators from the presented annotation study as input data for the described person identification approach.
For testing, we utilize further archive image data from DRA as well as additional images crawled from Google. After detecting faces and manually deleting irrelevant faces, we obtain a total of 350 face images which constitute our test set.
\subsubsection{Experiments}
Following the described approach, we independently build a training dictionary for each of the expert and non-expert study participants which rely on an annotator's personal judgments from the former annotation study and allow us to directly compare the person identification performance for different sets of annotators.
After evaluating classification accuracy based on each participant's training dictionary individually, we average the estimated performances for experts and non-experts, respectively. Furthermore, we generate ground truth labels considering the majority vote of the five experts and also evaluate the performance for this dictionary. For the non-experts we perform the majority vote based approach accordingly.
\subsection{Results}
\begin{table}[!bp]
  \caption{Person identification results by means of classification accuracies. Results are based on averaged individual expert E(1) and non-expert system NE(1) accuracies as well as on systems based on majority voting of respectively five expert E(5) and non-expert NE(5) annotations.}
  \label{tab:personid_results}
  \centering
  \begin{tabular}{l C{1.7cm} C{1.7cm} C{1.7cm} C{1.7cm}}
    \toprule
    Person & E(1) & NE(1) & E(5) & NE(5) \\
    \midrule
    Christa Wolf &  76.00~\% & 72.80~\% & 76.00~\% & 64.00~\% \\
    Erich Honecker &  87.13~\% & 85.94~\% & 86.14~\% & 87.13~\% \\
    Fritz Cremer & 88.33~\% & 83.33~\% & 91.67~\% & 83.33~\%  \\
    Hermann Henselmann &  53.33~\% & 48.33~\% & 75.00~\% & 66.67~\%\\
    Hilde Benjamin &  86.67~\% & 86.67~\% & 87.50~\% & 87.50~\% \\
    Michail Gorbatschow & 95.45~\% & 95.45~\% & 95.45~\% & 95.45~\% \\
    Siegmund Jaehn &  86.06~\% & 81.52~\% & 84.85~\% & 81.82~\% \\
    Stephan Hermlin &  64.12~\% & 60.29~\% & 70.59~\% & 67.65~\% \\
    Walter Ulbricht & 76.29~\% & 77.29~\% & 75.71~\% & 78.57~\%  \\
    Werner Tuebke & 82.35~\% & 81.76~\% & 82.35~\% & 82.35~\% \\
    \midrule
    Overall & 80.97~\% & 79.43~\% & 82.00~\% & 80.86~\% \\
    \bottomrule
\end{tabular}
\end{table}


%
Comparing the results of single expert and non-expert systems in Table~\ref{tab:personid_results}, expert annotations lead to an average increase of over $1.5~\%$ in performance. For only three of the ten personalities non-expert systems can keep up with those of experts. Only in the case of \emph{Walter Ulbricht} the performance is superior by $1~\%$. For the other seven personalities expert training images lead to an increase of up to $5~\%$ in accuracy.
In accordance with expectations, annotations determined using the majority vote of respectively five participants leads to an increase in performance for both experts and non-experts in comparison to individual judgments. The majority vote allows for more certain labels and acts as a filter for noise such as individual incorrect judgments or accidental false annotations. However, non-expert majority voting is still superior to that of single as well as multiple experts.
For \emph{Christa Wolf} the majority vote annotations even lead to a decrease of accuracy by almost $8.8~\%$ over the individual non-expert systems. This may be due to a minority of non-expert labels lifting the average for single non-expert systems, but which do not have an influence on majority voting.

As for the number of annotated images used per system, we cannot measure a significant correlation to the person identification results (average correlation: 0.297). Therefore, we assume that differences in performance are not caused by imbalanced training data but by incorrectly annotated person images.

Although the results suggest a moderate impact of labeling quality on person identification performance, we argue that in a real world annotation task, in which archive material can depict arbitrary persons, identifying more than ten persons is even harder. In this regard, labeling errors may have a higher impact on system performance. Please also note that non-expert participants were German students being more or less familiar with GDR history. Hence, labels of annotators who are completely unaware of German-German history could be more erroneous, presumably causing a higher performance gap between expert and non-expert person identification.
\section{Conclusions}
\label{conclusions}
In this paper, we have presented an inter-coder agreement study for historical image annotation involving both expert and non-expert annotators. We conducted our study for different image categories of 20 concepts and 10 personalities associated with the former GDR. In this regard, we evaluated whether amateur annotators can compete with the expertise of DRA archivists. We found that non-experts are as reliable annotators as experts for common sense and GDR-specific concepts. However, in identifying historical personalities, non-experts were more insecure which is reflected in a lower agreement. We also found that non-experts agree stronger with experts than with non-experts on personalities presumably due to a fraction of 'better' labelers among the non-experts.

In an attempt to identify the importance of inter-coder agreement for the performance of visual concept classification, we could determine correlations between AP results, training data size, and inter-coder agreement. Using support vector regression, we jointly modeled AP prediction for visual concepts by the amount of training data and agreement value. Our results indicate that it is very important to consider inter-coder agreement when designing a lexicon of visual concepts for a specific domain.

We also extended our study to determine how labeling quality affects system performance for the task of person identification. Single expert systems in average outperformed systems based on single non-expert annotation sets. In this context, we argued that differences in performance may even rise in a more complex real-world annotation task involving more persons to identify among even more unknown persons in the data. It turned out that majority voting corrects erroneous annotations and thus positively affects system performance. However, when it comes to annotating large-scale image material that current deep learning methods require, it is usually not plausible to annotate several times. Also, employing crowdworkers for co-annotation may not create as accurate labels as the German students considered non-experts in our case due to lack of domain knowledge.

\section*{Acknowledgements}
This work is financially supported by the German Research Foundation (DFG: Deutsche Forschungsgemeinschaft, project numbers: EW 134/4-1, FR 791/15-1, HO 5800).

%
%
%

\end{document}